\documentclass{article}

\usepackage[final]{neurips_2024}

\usepackage[utf8]{inputenc} 
\usepackage[T1]{fontenc}    
\usepackage{hyperref}       
\usepackage{url}            
\usepackage{booktabs}       
\usepackage{amsfonts}       
\usepackage{nicefrac}       
\usepackage{microtype}      
\usepackage{xcolor}         
\usepackage{enumitem}
\usepackage{amsmath}
\usepackage{graphicx}

\title{PhysVideoGenerator: Towards Physically Aware Video Generation via Latent Physics Guidance}

\author{
  Siddarth Nilol Kundur Satish\\
  Center for Data Science\\
  New York University\\
  \texttt{sk12590@nyu.edu} \\
   \And
   Devesh Jaiswal \\
   Department of Computer Science \\
   New York University \\
   \texttt{dj1380@nyu.edu} \\
   \AND
   Hongyu Chen \\
   Department of Computer Science \\
   New York University \\
   \texttt{hc4569@nyu.edu} \\
   \And
   Abhishek Bakshi \\
   Center for Data Science\\
  New York University\\
   \texttt{ab13573@nyu.edu} \\
}

\begin{document}

\maketitle

\begin{abstract}
Current video generation models produce high-quality aesthetic videos but often struggle to learn representations of real world physics dynamics, resulting in artifacts such as unnatural object collisions, inconsistent gravity, and temporal flickering. Existing solutions typically rely on massive scale or expensive external simulators, neither of which offers a tractable path for general-purpose, physically grounded generation. In this work, we propose PhysVideoGenerator, a proof-of-concept framework that explicitly embeds a learnable physics prior into the video generation process. We introduce a lightweight predictor network, PredictorP, which regresses high-level physical features extracted from a pre-trained Video Joint Embedding Predictive Architecture (V-JEPA 2)\cite{vjepa} directly from noisy diffusion latents. These predicted physics tokens are injected into the temporal attention layers of a DiT-based generator (Latte\cite{latte}) via a dedicated cross-attention mechanism. Our primary contribution is demonstrating the technical feasibility of this joint training paradigm: we show that (1) diffusion latents contain sufficient information to recover V-JEPA 2 physical representations, and (2) the multi-task optimization of diffusion and physics prediction losses remains stable over training. To enable scalable experimentation, we build a streaming dataset pipeline that converts raw videos into aligned latent, visual, and textual embeddings without storing intermediate video files. While our design is motivated by the belief that predictive world-model representations can improve physics awareness in video generation, we observe that jointly training predictive and diffusion components presents substantial optimization and memory challenges due to resource constraints. This report documents the architectural design, the technical challenges in aligning semantic and physical latent spaces, and the validation of the training stability, validation of training stability through convergence analysis, establishing a foundation for future large-scale evaluation of physics-aware generative models. The project repository is available at \url{https://github.com/CVFall2025-Project/PhysVideoGenerator}
\end{abstract}

\section{Introduction}

Recent progress in video diffusion models has led to impressive results in video generation due to the scaling of model capacity and training data. However, despite strong visual aesthetics, these models tend to lack structure for modeling the underlying physics dynamics of a scene, as they rely on implicit pattern learning from large datasets. This limitation can be seen in settings where physical consistency, temporal coherence, or long-horizon prediction is required. Therefore, there remains a gap between high-quality video generation and learning predictive representations that resemble world models. Addressing this limitation is critical for applications ranging from cinematic production to world simulation. Current solutions largely rely on two extremes: implicit scaling, which hopes physics emerges from massive datasets, and explicit simulation, which binds generation to rigid physics engines. Both have significant drawbacks: scaling is resource-intensive, while simulators struggle to generalize to "in-the-wild" scenes.

Joint Embedding Predictive Architectures (JEPAs)\cite{ijepa} have emerged as a promising framework for learning abstract, predictive representations by training models to forecast future latent embeddings rather than reconstruct raw observations on the pixel level. Such representations have shown potential for capturing semantic and dynamical structure while avoiding pixel-level shortcuts. Motivated by this perspective, we investigate whether VJEPA 2-style predictive signals can serve as an inductive bias for video diffusion models, guiding generation toward more structured and physically consistent dynamics without sacrificing expressive power. 

In the early stages of this project, we explored implementing our framework on top of the CogVideoX-2B\cite{cogvideox} backbone due to its strong performance in large-scale text-to-video generation. However, the combination of CogVideoX’s model size, high-dimensional latent representations, and additional physics-conditioning modules introduced prohibitive memory overhead during both training and inference, resulting in persistent out-of-memory failures namely since we only have access to at most two A100 NVIDIA GPUs. To enable systematic experimentation and isolate the effects of latent physics guidance under realistic resource constraints, we transitioned to the Latte-1 architecture. Latte’s factorized spatial–temporal attention design provides a more memory-efficient backbone while preserving the expressiveness required for physics-aware conditioning, making it a more suitable platform for prototyping and validating our approach.

\subsubsection*{1.1 Project Goals and Contributions}
The primary objective of this work was to design, implement, and validate the feasibility of a framework that conditions a video generator (Latte) on physical priors. Our specific contributions are:

\begin{enumerate}[leftmargin=*, nosep, label=\arabic*.]
    \item \textbf{Architectural Design of PhysVideoGenerator:} We propose a novel architecture that integrates a ``physics predictor'' branch into a DiT-based video generator. This branch learns to regress V-JEPA 2 embeddings from noisy latents, effectively forcing the model to ``understand'' the physics of the scene.
    
    \item \textbf{Implementation of Joint Training Pipeline:} We successfully engineered a training loop that optimizes the diffusion loss and the physics prediction loss simultaneously. This required solving complex technical challenges related to tensor expansion and memory management on high-performance compute nodes (A100).
    
    \item \textbf{Feasibility Validation:} We validated the training stability of this multi-task learning objective over 50 training epochs, demonstrating that the physics loss converges alongside the diffusion noise loss. This confirms that diffusion latents contain sufficient information to recover physical representations and establishes the viability of our approach for future large-scale experimentation.
\end{enumerate}

While full-scale benchmarking on datasets like VideoPhy-2\cite{videophy2} remains future work, this report establishes the methodological foundation and technical feasibility of Latent Physics Guidance as a scalable direction for physically consistent video generation.


\section{Related Works}
Recent advances in generative modeling have largely shifted from U-Net-based architectures to Transformer-based backbones, pioneered by the Diffusion Transformer (DiT)\cite{dit}. Peebles and Xie\cite{dit} demonstrated that the inductive bias of U-Nets is not strictly necessary for diffusion and that Transformers, with their scalability and global attention mechanisms, can achieve state-of-the-art results when treated as standard sequence modeling problems. Building on this foundation, Latte\cite{latte} extended the DiT architecture to the temporal domain. Ma et al. proposed a factorized attention mechanism—separating spatial and temporal processing—to efficiently model video dynamics. By applying the DiT design to video latent spaces, Latte achieved high-fidelity results with significantly lower computational complexity than full 3D-attention models. Our work adopts Latte as the generative backbone, leveraging its factorized design to inject physical priors specifically into the temporal evolution blocks.

While architectural improvements like Latte enhance visual fidelity, ensuring temporal and semantic consistency remains a challenge. A promising direction to address this is "Representation Alignment," as explored in Video-REPA\cite{videorepa}. This line of work posits that diffusion models can be guided by distilling features from powerful, pre-trained discriminative models (such as DINO or JEPA) directly into the generative process. Video-REPA specifically demonstrated that enforcing consistency between diffusion features and pre-trained video representations can significantly improve temporal coherence. Our approach, PhysVideoGenerator, is directly motivated by this insight. However, instead of general representation alignment, we specifically target physical alignment. By aligning our diffusion latents with V-JEPA 2 - a model trained explicitly on predictive world modeling, we move beyond simple temporal smoothness to enforce complex physical commonsense, such as gravity, object permanence, and realistic interaction dynamics.

\section{Materials and Methods}
\subsection{Experimental Setup}
\paragraph{Dataset.}
We use the OpenVid-1M\cite{openvid1m} dataset for our experiments, a large-scale collection of high-quality video-text pairs. OpenVid-1M is chosen for its diversity in motion patterns and high visual fidelity, which provides a robust foundation for learning generalizable physical dynamics. We process video clips at a resolution of 256×256 with a temporal duration of 16 frames.

\paragraph{Feature Extraction Pipeline.}
To decouple feature extraction from the generative training loop, we pre-compute and cache latent representations for three modalities. The specific configurations are as follows:

\begin{itemize}[leftmargin=*, nosep]
    \item \textbf{Variational Autoencoder Latents (VAE):} We employ the pre-trained Variational Autoencoder from the \textbf{Latte-1} framework. Input videos $x \in \mathbb{R}^{3 \times 16 \times 256 \times 256}$ are encoded into compressed latents $z_0 \in \mathbb{R}^{4 \times 16 \times 32 \times 32}$, corresponding to a spatial downsampling factor of 8.\vspace{2mm}
    
    \item \textbf{Text Embeddings:} Semantic conditioning is derived from the \textbf{T5-v1.1-XXL} encoder. Text prompts are tokenized to a fixed sequence length of $L=226$, resulting in embeddings $c_{\text{text}} \in \mathbb{R}^{226 \times 4096}$. These embeddings condition the spatial transformer blocks. \vspace{2mm}
    
    \item \textbf{Physical Priors (V-JEPA 2):} To capture physical commonsense, we utilize the \textbf{V-JEPA 2} (Video Joint Embedding Predictive Architecture) pre-trained model. Unlike semantic encoders (e.g., CLIP), V-JEPA 2 is optimized for predictive modeling of temporal dynamics. We extract patch-level representations from the final encoder layer, yielding physics tokens $c_{\text{phys}} \in \mathbb{R}^{2048 \times 1408}$, where 2048 corresponds to the flattened spatiotemporal patches ($16 \times 16$ spatial $\times$ 8 temporal).
\end{itemize}

\paragraph{Implementation Details.}
We implement our framework using PyTorch and the Diffusers library. The backbone is initialized with \textbf{Latte-1} weights. We adopt a fine-tuning strategy where we freeze the VAE, Text Encoder, and the original Spatial Transformer blocks. We explicitly train the \textit{PredictorP} module (a 4-layer lightweight Transformer with hidden dimension $d=512$) and the newly introduced Temporal Cross-Attention layers. 

Optimization is performed using AdamW with a learning rate of $1e-5$ and weight decay of $0.01$. To mitigate the memory constraints of 3D attention on high-dimensional V-JEPA 2 tokens, we utilize Gradient Checkpointing and Mixed Precision (BF16) on 1 NVIDIA A100 (40GB) GPU provided on Google Cloud Burst.

\subsection{Architecture}

Our framework builds upon the \textbf{Latte-1} Latent Diffusion Transformer based Text-to-Video (T2V) Generation model, modifying it to enable latent physics guidance. The architecture consists of two primary components: the \textbf{PredictorP} network, which infers physical dynamics from noisy latents, and the \textbf{Physics-Conditioned Temporal Blocks}, which inject these priors into the generation process.

\begin{figure}[h]
    \centering
    \includegraphics[width=0.7\textwidth]{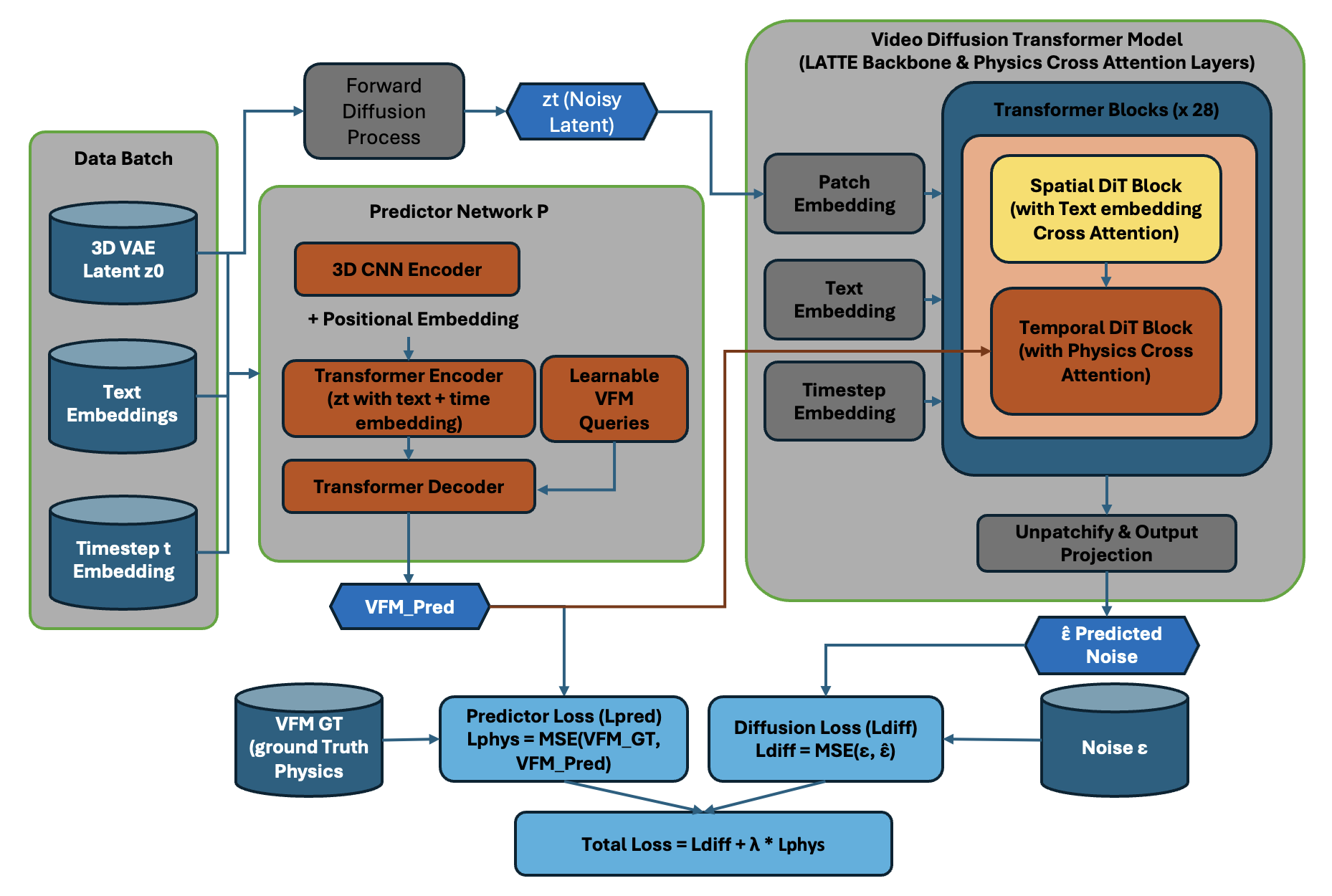}
    \caption{PhysVideoGenerator Architecture}
    \label{fig:example}
\end{figure}

\subsubsection{The Physics Predictor Network (PredictorP)}
A core challenge in physically grounded generation is that ground-truth physical descriptors are unavailable during inference. To address this, we introduce \textbf{PredictorP}, an auxiliary network that learns to regress the physical representation $p_{target}$ directly from the noisy diffusion state $z_t$, text embedding $c_{\text{text}}$, and timestep embedding $t_{emb}$. PredictorP operates in three stages:

\begin{enumerate}[leftmargin=*]
    \item \textbf{Latent Encoding:} The input noisy latents $z_t$ are first processed by a lightweight 3D Convolutional encoder to capture local spatiotemporal features:
    \begin{equation}
    h_{vis} = \text{Conv3DEnc}(z_t) \quad \in \mathbb{R}^{d \times (T/2) \times (H/2) \times (W/2)}
    \end{equation}
    
    \item \textbf{Multi-Modal Fusion:} The encoded visual features are flattened and concatenated with the text embeddings $c_{text}$ and the diffusion timestep embedding $t_{emb}$. This fused sequence is processed by a Transformer Encoder:
    \begin{equation}
    h_{fused} = \text{TransEnc}([h_{vis}; c_{text}; t_{emb}])
    \end{equation}
    
    \item \textbf{V-JEPA 2 Decoding:} We employ a Transformer Decoder with \textbf{learnable queries} $Q_{phys} \in \mathbb{R}^{N \times d}$ ($N=2048$). These queries cross-attend to $h_{fused}$ to generate the predicted physical tokens $\hat{p}$:
    \begin{equation}
    \hat{p} = \text{Linear}(\text{TransDec}(Q_{phys}, h_{fused}))
    \end{equation}
\end{enumerate}

\subsubsection{Temporal Physics-Cross-Attention}
To inject the predicted physics $\hat{p}$ into the generative process, we modify the standard Temporal Transformer Blocks. We introduce a \textbf{Physics-Cross-Attention} layer immediately following the temporal self-attention.

Let $x_{temp} \in \mathbb{R}^{F \times d}$ be the hidden states of a specific spatial patch across $F$ frames. We formulate the injection as:
\begin{equation}
x'_{temp} = x_{temp} + \text{Attention}(Q=W_q x_{temp}, K=W_k \hat{p}, V=W_v \hat{p})
\end{equation}
By broadcasting the physics tokens, we ensure that the temporal evolution of every patch is globally consistent with the predicted physical laws.

\subsubsection{Joint Training Objective}
We train the PredictorP and the diffusion temporal layers simultaneously. The total objective is a weighted sum of the diffusion noise prediction loss and the physics regression loss:
\begin{equation}
\mathcal{L}_{total} = \underbrace{||\epsilon - \epsilon_\theta(z_t, t, c_{text}, \hat{p})||^2}_{\text{Diffusion Loss}} + \lambda \cdot \underbrace{||\hat{p} - p_{gt}||^2}_{\text{Physics Loss}}
\end{equation}
Where $p_{gt}$ are the ground-truth V-JEPA 2 tokens. This encourages the diffusion backbone to actively utilize the physics tokens for denoising.

\subsection{Inference Methodology}

Our inference pipeline generates videos conditioned on text prompts by running reverse diffusion in the latent space of a pretrained Latte VAE. For each prompt, we compute token-level text embeddings using a T5 encoder (we use \texttt{google/t5-v1\_1-xxl} and a fixed maximum sequence length to match the training interface), initialize a Gaussian latent tensor
\[
z_T \sim \mathcal{N}(0, I),
\]
with the spatiotemporal latent shape used during training, and iteratively denoise it for $T$ steps using an $\epsilon$-prediction DDPM sampler. At each reverse diffusion step $t$, we compute a sinusoidal timestep embedding and run the learned predictor $P$ on the current latent $z_t$, the text embeddings, and the timestep embedding to produce a sequence of ``physics'' tokens. These predicted tokens are injected into the generator via physics cross-attention, where a learned gate controls the strength of physics conditioning. The denoiser outputs $\hat{\epsilon}_t$, which we use to form the DDPM posterior mean update and sample $z_{t-1}$ (adding Gaussian noise for $t>0$). After completing the reverse process, we obtain a final latent $z_0$, rescale by the VAE scaling factor, decode into pixel space, convert outputs from $[-1,1]$ to $[0,255]$, and save as video.

For debugging and systems validation, we retain a "dry-run" mode that skips checkpoint loading and runs a single denoising step on a single prompt to verify end-to-end execution without the full sampling cost.

\section{Training Feasibility Analysis}
The central empirical question of this work is whether the proposed joint training paradigm is viable: specifically, whether (1) the physics prediction loss can be optimized alongside the diffusion loss without destabilizing training, and (2) the diffusion latents contain sufficient information to regress V-JEPA 2 representations.

\subsection{Training Configuration}
We trained PhysVideoGenerator for 50 epochs on a subset of OpenVid-1M using a single NVIDIA A100 (40GB) GPU. The physics loss weight was set to $\lambda = 0.1$ after preliminary experiments showed this provided a balanced gradient signal between the two objectives.

\subsection{Convergence Analysis}
\begin{figure}[h]
    \centering
    \includegraphics[width=0.7\textwidth]{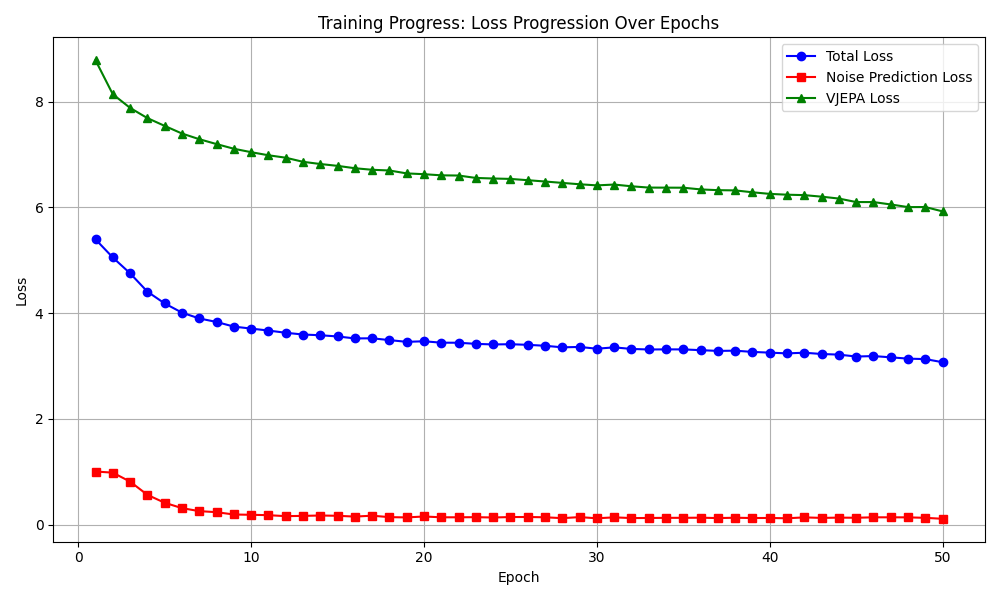}
    \caption{Training Convergence}
    \label{fig:training}
\end{figure}

We monitored both loss components as shown in Figure\ref{fig:training} throughout training to assess optimization stability.

\paragraph{Diffusion Loss ($\mathcal{L}_{diff}$).} The noise prediction loss followed a typical diffusion training trajectory, decreasing steadily over the first 10 epochs before entering a slower convergence regime. This indicates that the addition of the physics conditioning branch does not fundamentally disrupt the diffusion learning dynamics.

\paragraph{Physics Prediction Loss ($\mathcal{L}_{phys}$).} The physics regression loss exhibited consistent decrease throughout training, demonstrating that the PredictorP network successfully learns to map noisy diffusion latents to V-JEPA 2 representations. Notably, the physics loss converged faster than the diffusion loss in early training, suggesting that the latent-to-physics mapping is learnable even from partially denoised representations.

\paragraph{Key Finding:} The simultaneous decrease of both losses without oscillation or divergence validates that the multi-task objective is well-posed. This is a non-trivial result, as jointly optimizing generative and predictive objectives can lead to competing gradients and unstable training dynamics.

\subsection{Memory and Computational Considerations}
A significant engineering challenge was managing GPU memory given the high dimensionality of V-JEPA 2 tokens (2048 × 1408). We successfully mitigated memory pressure through:

\begin{enumerate}[leftmargin=*]
    \item \textbf{Gradient Checkpointing} Recomputing activations during the backward pass reduced peak memory by approximately 40\%.
    
    \item \textbf{Mixed Precision Training (BF16)} Reduced memory footprint while maintaining numerical stability.
    
    \item \textbf{Frozen Components} Freezing the VAE, text encoder, and spatial attention blocks focused trainable parameters on the physics-relevant modules.
    
\end{enumerate}

These optimizations enabled training on a two 40GB A100 GPUs, demonstrating that the approach is accessible to academic compute budgets.

\subsection{Limitations of Current Validation}
While our training analysis demonstrates optimization feasibility, several limitations remain:
\begin{enumerate}[leftmargin=*]
    \item \textbf{No Generative Quality Evaluation} We did not conduct inference to assess whether the trained model produces physically plausible videos. This requires substantial additional compute and systematic evaluation.
    
    \item \textbf{Limited Training Scale} 50 epochs on a data subset is insufficient to fully train either the predictor or the adapted diffusion layers. Large-scale training is necessary to assess the method's true potential.
    
    \item \textbf{No Ablation Studies} We did not compare against baselines or ablated variants, which would be necessary to attribute any quality improvements to specific architectural choices.
    
\end{enumerate}

\section{Proposed Evaluation Protocol}
We outline below the evaluation protocol for future work to assess temporal coherence, semantic alignment, and physical plausibility of generated videos. These metrics are standard in contemporary video generation and physical reasoning benchmarks and were selected to reflect complementary aspects of video quality.

\paragraph{Optical Flow Consistency.}
To assess low-level temporal smoothness, we planned to measure dense optical flow consistency across consecutive frames using Farnebäck’s algorithm. Optical flow–based metrics are commonly used to detect flickering, jitter, and unstable object boundaries that may not be captured by semantic similarity measures. Average flow magnitude and its temporal variation were intended to serve as indicators of frame-to-frame stability.

\paragraph{Motion Consistency in Representation Space.}
To evaluate whether motion remains semantically coherent over time, we intended to measure temporal consistency in a learned representation space using VideoCLIP\cite{videoclip} embeddings. Specifically, cosine similarity between embeddings of consecutive frames was used as a proxy for whether object identity and high-level scene semantics remain stable across time, helping to distinguish coherent motion from visually smooth but semantically inconsistent artifacts.

\paragraph{Perceptual Temporal Consistency (T-LPIPS).}
We further planned to evaluate perceptual smoothness using Temporal LPIPS (T-LPIPS)\cite{lpips} with a VGG\cite{vgg}-based backbone. T-LPIPS measures perceptual differences between consecutive frames in a deep feature space and is widely used to detect temporal artifacts that are perceptually salient but difficult to capture with pixel-level metrics.

\paragraph{VideoPhy Semantic Adherence.}
Semantic alignment between generated videos and conditioning text prompts was intended to be evaluated using the VideoPhy\cite{videophy2} semantic adherence metric. This benchmark measures video–text alignment using pretrained video–language models and reflects whether the generated content corresponds to the intended actions and objects described in the prompt, independent of physical realism.

\paragraph{VideoPhy Physical Correctness.}
To assess physical plausibility, we planned to use the VideoPhy physical correctness metric, which evaluates violations of intuitive physics such as implausible motion, gravity inconsistencies, and non-causal interactions. This metric is designed to capture commonsense physical reasoning rather than exact numerical accuracy.

\paragraph{VideoPhy-2 Evaluation.}
Finally, we intended to report semantic adherence and physical correctness scores on VideoPhy-2\cite{videophy2}, a more challenging extension that emphasizes longer temporal horizons and complex interactions. VideoPhy-2 is particularly relevant for evaluating whether physics-aware conditioning improves sustained physical consistency over time.

Overall, this evaluation suite was chosen to reflect a separation of concerns: optical flow and perceptual metrics for low-level temporal stability, representation-based metrics for semantic motion coherence, and benchmark-based metrics for prompt alignment and physical plausibility. While full quantitative evaluation remains future work, this protocol provides a path for systematically assessing physics-aware video generation.

\section{Future Work}

\subsection{Planned Experiments}
\paragraph{Ablation Studies} We plan to conduct systematic ablations comparing:

\begin{enumerate}[leftmargin=*]
    \item \textbf{Baseline} The original Latte backbone without physics injection.
    
    \item \textbf{w/o Physics Attn} PredictorP trained but cross-attention disabled.
    
    \item \textbf{w/o LoRA} Full model with completely frozen backbone.
    
\end{enumerate}

\paragraph{Comparative Analysis} We propose comparison against state-of-the-art open-source video generation models (OpenSora, VideoCrafter2, HunyuanVideo) on prompts designed to test physical interactions.

\subsection{Technical Extensions}
\begin{enumerate}[leftmargin=*]
    \item \textbf{Efficient Physics Token Representations} The current 2048×1408 V-JEPA 2 representation is memory-intensive. Investigating compressed or pooled representations could improve scalability.
    
    \item \textbf{Alternative Backbones} While we transitioned from CogVideoX to Latte due to memory constraints, future work with larger compute budgets could explore physics conditioning on more capable backbones.
    
    \item \textbf{Inference-Time Physics Guidance} Exploring classifier-free guidance variants that modulate physics conditioning strength during inference.
    
\end{enumerate}

\section{Conclusion}

In this work, we presented \textit{PhysVideoGenerator}, a proof-of-concept framework for incorporating physics-aware predictive representations into a latent video diffusion model built on the Latte-1 architecture. By augmenting a DiT-based video generator with a lightweight physics predictor trained to regress V-JEPA 2 representations directly from noisy diffusion latents, we demonstrated a principled mechanism for latent physics guidance that does not require external simulators or physical supervision at inference time. Leveraging Latte’s factorized spatial–temporal attention structure, our approach injects predicted physics tokens specifically into the temporal evolution blocks, enabling targeted conditioning of motion dynamics while preserving the model’s generative capacity.

The central contribution of this work is the validation of training feasibility: we showed that diffusion latents produced by Latte retain sufficient information to recover high-level physical representations, and that multi-task training of diffusion and physics prediction objectives remains stable over 50 training epochs. This establishes the technical viability of our approach, even though comprehensive evaluation of generative quality remains future work.
Beyond architectural design, we successfully engineered a joint training pipeline that optimizes diffusion denoising and physics prediction objectives simultaneously within the Latte framework. Our solutions for memory management—including gradient checkpointing, mixed precision training, and strategic component freezing—demonstrate that the approach is tractable on academic compute budgets.
These results highlight both the promise and the challenges of integrating world-model representations into video generators. While we have not yet demonstrated improved video quality, the stability of joint training and the learnability of the latent-to-physics mapping provide a foundation for future large-scale experimentation. We hope this work motivates continued research on more efficient physics token representations, improved optimization strategies, and tighter coupling between predictive and generative components.

{
    \small
    \bibliographystyle{plain} 
    \bibliography{references}

@inproceedings{latte,
  title={Latte: Latent Diffusion Transformer for Video Generation},
  author={Ma, Xin and Wang, Yaohui and Jia, Gengyun and Chen, Xinyuan and Liu, Ron and Li, Yu-Gang and Chen, Cilin and Qiao, Yu},
  booktitle={Proceedings of the IEEE/CVF Conference on Computer Vision and Pattern Recognition (CVPR)},
  year={2024}
}

@inproceedings{ijepa,
  title={Self-Supervised Learning from Images with a Joint-Embedding Predictive Architecture},
  author={Assran, Mahmoud and Duval, Quentin and Misra, Ishan and Bojanowski, Piotr and Vincent, Pascal and Rabbat, Michael and LeCun, Yann and Ballas, Nicolas},
  booktitle={Proceedings of the IEEE/CVF Conference on Computer Vision and Pattern Recognition (CVPR)},
  year={2023}
}

@article{vjepa,
  title={V-JEPA 2: Self-Supervised Video Models Enable Understanding, Prediction and Planning},
  author={Assran, Mahmoud and Bardes, Adrien and Fan, David and Garrido, Quentin and Howes, Russell and Komeili, Mojtaba and Muckley, Matthew and Rizvi, Ammar and Roberts, Claire and Sinha, Koustuv and others},
  journal={arXiv preprint arXiv:2506.09985},
  year={2025}
}

@inproceedings{dit,
  title={Scalable Diffusion Models with Transformers},
  author={Peebles, William and Xie, Saining},
  booktitle={Proceedings of the IEEE/CVF International Conference on Computer Vision (ICCV)},
  year={2023}
}

@article{openvid1m,
  title={OpenVid-1M: A Large-Scale High-Quality Dataset for Text-to-Video Generation},
  author={Nan, Haowei and Xie, Chi and Yin, Xi and others},
  journal={arXiv preprint arXiv:2407.02371},
  year={2024}
}

@article{videorepa,
  title={Video-REPA: Temporal Representation Alignment for Video Diffusion},
  author={Zhang, Yanzhe and others},
  journal={arXiv preprint},
  year={2024},
  note={Preprint}
}

@article{videophy2,
  title={VideoPhy-2: A Challenging Action-Centric Physical Commonsense Evaluation in Video Generation},
  author={Bansal, Hritik and Peng, Clark and Bitton, Yonatan and Goldenberg, Roman and Grover, Aditya and Chang, Kai-Wei},
  journal={arXiv preprint arXiv:2503.06800},
  year={2025}
}

@article{videoclip,
  title={VideoCLIP: Contrastive Pre-training for Zero-shot Video-Text Understanding},
  author={Xu, Hu and Das, Abir and Wang, Jason and et al.},
  journal={arXiv preprint arXiv:2109.14084},
  year={2021}
}

@article{lpips,
  title={The Unreasonable Effectiveness of Deep Features as a Perceptual Metric},
  author={Zhang, Richard and Isola, Phillip and Efros, Alexei A. and Shechtman, Eli and Wang, Oliver},
  journal={arXiv preprint arXiv:1801.03924},
  year={2018}
}

@article{vgg,
  title={Very Deep Convolutional Networks for Large-Scale Image Recognition},
  author={Simonyan, Karen and Zisserman, Andrew},
  journal={arXiv preprint arXiv:1409.1556},
  year={2014}
}

@article{cogvideox,
  title={CogVideoX: Text-to-Video Diffusion Models with An Expert Transformer},
  author={Yang, Zhuoyi and Teng, Jiayan and Zheng, Wendi and Ding, Ming and others},
  journal={arXiv preprint arXiv:2408.06072},
  year={2024}
}
}

\medskip

{
\small
}






\newpage

\end{document}